\documentclass[10pt,twocolumn,letterpaper]{article}

\usepackage{wacv}
\usepackage{times}
\usepackage{epsfig}
\usepackage{graphicx}
\usepackage{amsmath}
\usepackage{amssymb}
\usepackage{multirow}
\usepackage{makecell}
\usepackage{xcolor}
\usepackage{authblk}

\def\wacvPaperID{1229} 

\wacvfinalcopy 

\ifwacvfinal
\def\assignedStartPage{1} 
\fi

\ifwacvfinal
\usepackage[breaklinks=true,bookmarks=false]{hyperref}
\else
\usepackage[pagebackref=true,breaklinks=true,colorlinks,bookmarks=false]{hyperref}
\fi

\ifwacvfinal
\else
\fi

\begin{document}

\title{DeepMark++: Real-time Clothing Detection at the Edge}

\author[ ]{
	\normalsize Alexey Sidnev\textsuperscript{1,2},
	Alexander Krapivin\textsuperscript{1},
	Alexey Trushkov\textsuperscript{1},
    Ekaterina Krasikova\textsuperscript{1},
	Maxim Kazakov\textsuperscript{1,3},
    Mikhail Viryasov\textsuperscript{1}
}
\affil[1]{\small Huawei Research Center, Nizhny Novgorod, Russia}
\affil[2]{\small Lobachevsky State University of Nizhny Novgorod, Russia}
\affil[3]{\small National Research University Higher School of Economics, Nizhny Novgorod, Russia}
\affil[ ]{\texttt{\scriptsize {\{sidnev.alexey, krapivin.alexander, trushkov.alexey, krasikova.ekaterina, kazakov.maxim, viryasov.mikhail\}@huawei.com}}}

\maketitle

\begin{abstract}
   Clothing recognition is the most fundamental AI application challenge within the fashion domain.
   While existing solutions offer decent recognition accuracy, they are generally slow and require significant computational resources.
   In this paper we propose a single-stage approach to overcome this obstacle and deliver rapid clothing detection and keypoint estimation.
   Our solution is based on a multi-target network CenterNet~\cite{CenterNet}, and we introduce
   several powerful post-processing techniques to enhance performance.
   Our most accurate model achieves results comparable to state-of-the-art solutions on the DeepFashion2
   dataset~\cite{DeepFashion2}, and our light and fast model runs at 17 FPS on the Huawei P40 Pro smartphone.
   In addition, we achieved second place in the DeepFashion2 Landmark Estimation Challenge
   2020\protect\footnotemark ~with 0.582~$\mathit{mAP}$ on the test dataset.
\end{abstract}

\section{Introduction}\label{sec:introduction}

\footnotetext{\url{https://sites.google.com/view/cvcreative2020/deepfashion2}}

Image recognition is often the first step towards processing an image in many applications, and
involves locating, classifying and, sometimes, identifying the objects presented in the image.
An object's location can be estimated with different levels of
precision (bounding boxes, keypoints, or segmentation masks), and each introduces a new challenge.

\begin{figure}[ht]
	\centering
	\includegraphics[width=\columnwidth]{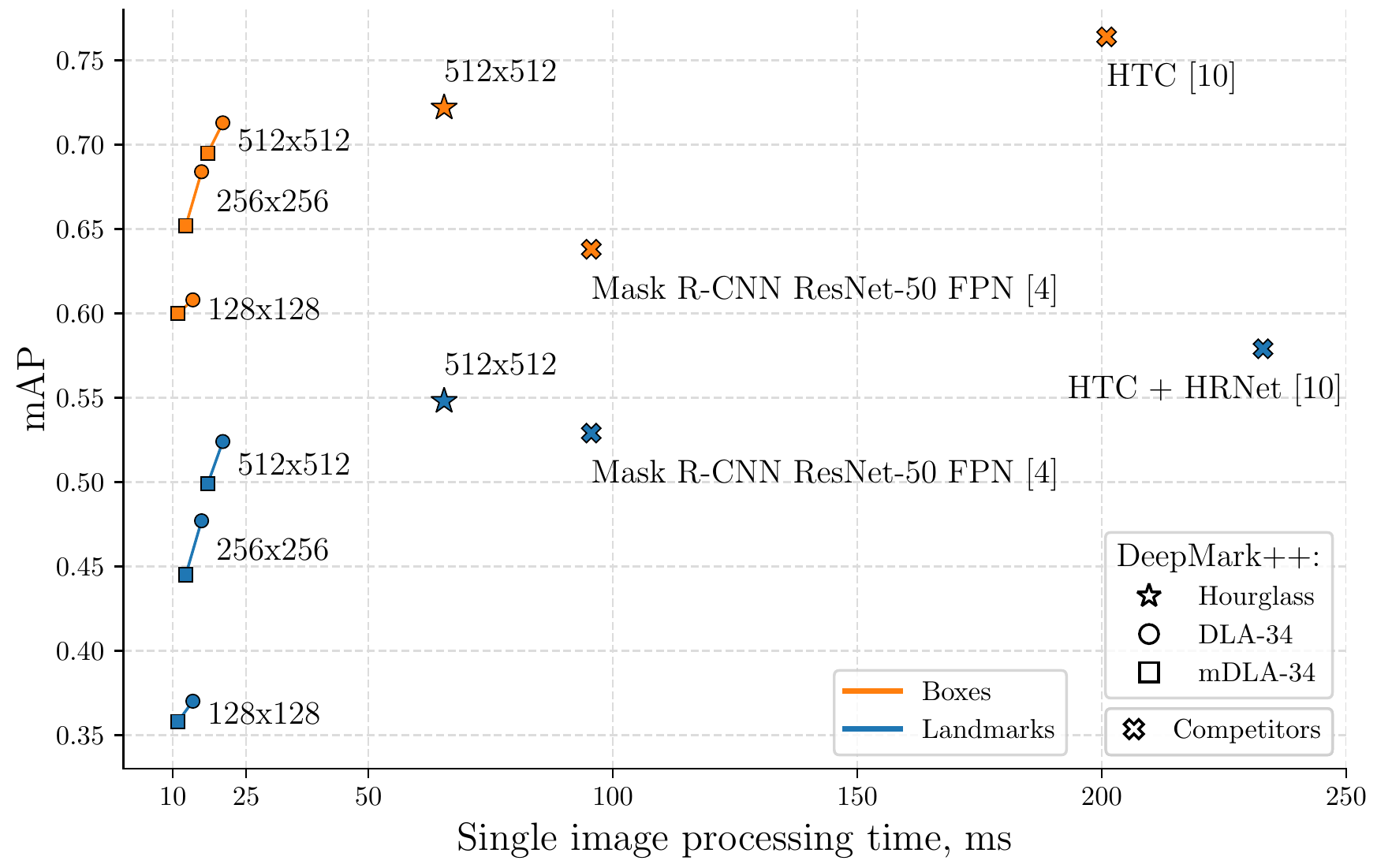}
	\caption{Speed/accuracy trade-off for object detection and landmark estimation
	on the DeepFashion2 validation dataset~\protect\cite{DeepFashion2} using an RTX 2080ti.
	Mask R-CNN~\protect\cite{DeepFashion2} time is estimated with Keypoint R-CNN model from Detectron2\protect\footnotemark[2].
	HTC~\cite{Lin2020AggregationAF} detection time is estimated with COCO detection model\protect\footnotemark[3].
	HTC~+~HRNet~\cite{Lin2020AggregationAF} is used without test-time augmentations in the single (aggregated) configuration.
		}
	\label{fig:plot}
\end{figure}

\footnotetext[2]{\url{https://github.com/facebookresearch/detectron2}}
\footnotetext[3]{\url{https://github.com/open-mmlab/mmdetection}}

Object detection aims to locate an object by estimating its boundaries.
A detected object is typically presented by a tight bounding box and a predicted class label.
A bounding box can be used directly in applications such as pedestrian detection~\cite{Zhang_2018_ECCV},
or to crop an object from the original image for further processing.

Keypoint detection aims to estimate the precise location of an object's keypoints, which are sometimes referred to as landmarks.
Given a set of correctly defined keypoints, one can transform an object in various ways -- for example, position alignment for future processing.
However, details of keypoint locations are not enough to estimate an object's
boundaries due to a variety of object forms and its orientation in space.
Instead, object detection and keypoint estimation are combined to provide enough
information to locate and process objects.

The performance of models that operate on keypoints depends heavily on the number of
unique keypoints defined in the task.
The Deepfashion2 dataset~\cite{DeepFashion2}, one of the largest and most diverse
datasets in the fashion domain, provides annotation for 13 classes of clothing, each
characterized by a certain set of keypoints (294 keypoints in total).

State-of-the-art methods used to solve keypoint estimation problems on the
DeepFashion2 dataset rely on heavy architectures such as Mask R-CNN~\cite{DeepFashion2} or
HTC~+~HRNet~\cite{Lin2020AggregationAF}, which require large amounts of GPU resources and extended training time,
and can not be considered to run on mobile devices.
The recently proposed DeepMark~\cite{DeepMark} uses a one-stage anchor-free detection model
called CenterNet~\cite{CenterNet} to simultaneously detect clothing items and estimate clothing landmark locations.
While this approach is fast in the inference phase, it is still both time- and memory-consuming at the training stage.
The massive number of unique keypoints in the DeepFashion2 dataset presents a significant bottleneck.

In this paper we propose a keypoint grouping strategy to drastically reduce the number of
keypoints and to accelerate the training process.
In addition, we propose and thoroughly study several post-processing techniques which can improve
the model's accuracy without compromising its performance.
Through this approach, we will demonstrate that our model is capable of achieving results comparable to the state-of-the-art.

Finally, we propose a light and fast version of our model capable of running on a mobile device.
We will demonstrate optimal real-time performance of 50 FPS and near real-time performance of 17 FPS
with an acceptable accuracy of 0.445 mAP for the keypoint estimation problem.

\section{Related work}\label{sec:related-work}
Object keypoints can typically be utilized in numerous applications.
For example, they can be used to identify a human pose~\cite{alej2016stacked}, locate facial
landmarks~\cite{zhang2014facial}, or to estimate 3D or 6D object locations~\cite{He_2020_CVPR}.
However, the application we are most concerned with in this paper is clothing landmark estimation.

\begin{figure}[t]
	\centering
	\includegraphics[width=\columnwidth]{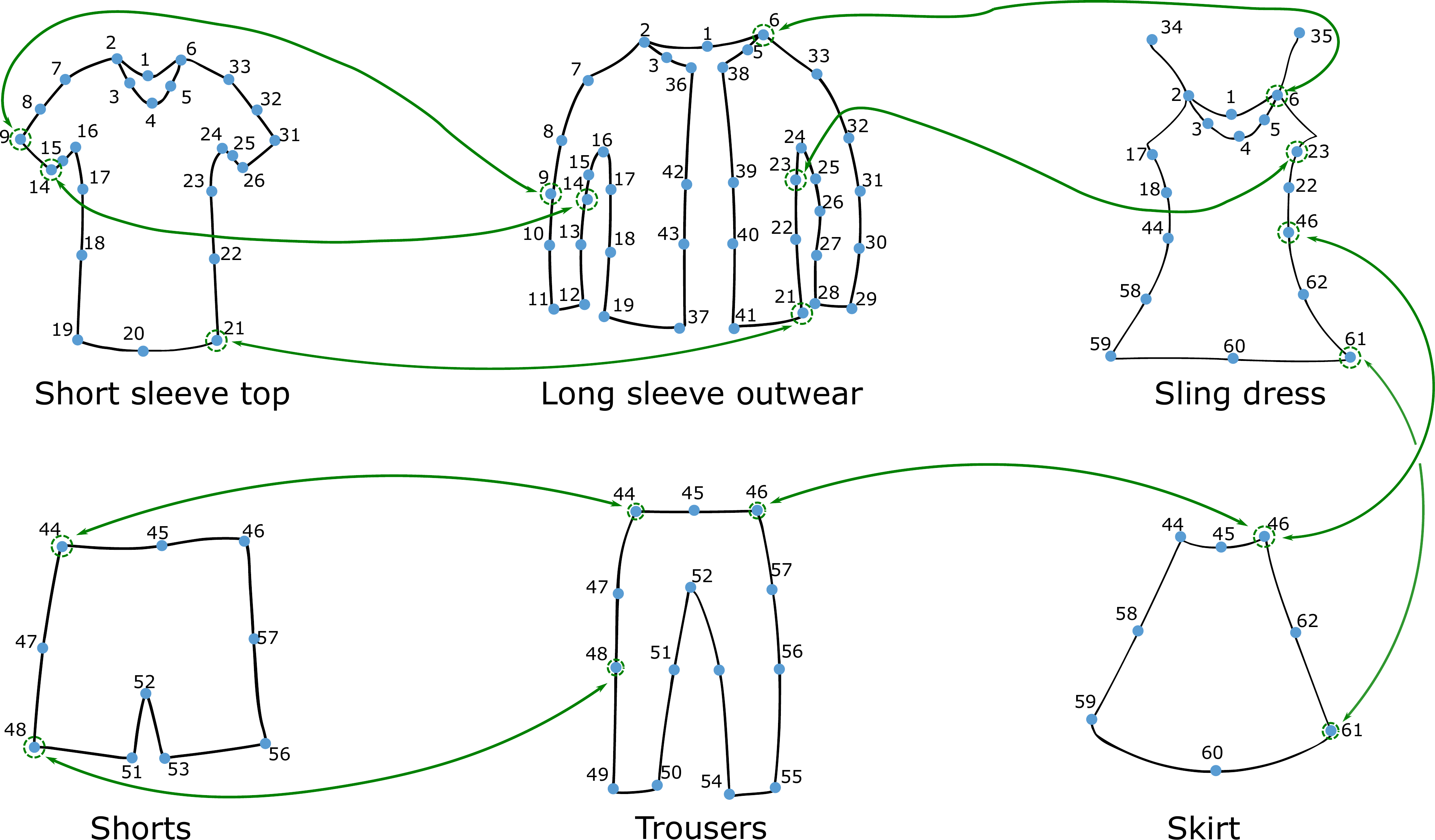}
	\caption{The semantic grouping approach.
	A number indicates the group a keypoint belongs to, and the green arrows are examples of merged keypoints.}
	\label{fig:keypoints}
\end{figure}

There are two open datasets capable of providing clothing images annotated with bounding boxes and keypoints.
DeepFashion~\cite{liu2016deepfashion} defines between 4 and 8 keypoints
for three classes of clothing items - top, bottom and full-body.
Apart from a modest variety of clothing items and sparse keypoint definition, DeepFashion is limited
to one object per image, making it less suitable for real-life applications.
DeepFashion2~\cite{DeepFashion2} improves on some of these drawbacks
by providing annotations for 13 classes, each characterized by a unique set of keypoints, from
8 to 39, for 294 keypoints in total.
Examples of keypoint definition are presented in Figure~\ref{fig:keypoints}.
The number of objects on an image ranges from 1 to 5.

Several papers are aiming to solve the landmark estimation challenge with regards to the DeepFashion2 dataset,
with the original solution proposed in~\cite{DeepFashion2} built upon the Mask R-CNN architecture~\cite{MaskRCNN}.
This solution uses the ResNet-FPN~\cite{lin2017feature} backbone and RoIAlign to extract features from different levels of the pyramid map,
which are then processed by two different branches, one each for
object detection and landmark estimation, in order to obtain final outputs.

In~\cite{Chen_2019_ICCV}, authors introduce an attention module to aggregate features from
different levels and to produce an output heatmap for landmark estimation.
While this approach has delivered an improvement of 0.02 mAP for landmark estimation, it does not
provide bounding boxes and instead offers only keypoint locations.

A recently proposed solution by Tzu-Heng Lin~\cite{Lin2020AggregationAF} offers current
state-of-the-art capabilities with 0.614 mAP for landmark estimation and 0.764 mAP for object detection.
This top-down approach first uses Hybrid Task Cascade~\cite{chen2019hybrid}
to perform object detection and obtain bounding boxes, and then runs
HRNet-w48~\cite{sun2019deep} on a cropped image to estimate keypoint locations.
This method also uses the aggregation strategy to reduce the number of keypoints and finetune
the landmark estimation model for each class separately, thereby enhancing performance.

As existing solutions rely on slow and resource-consuming architecture,
they are unsuitable for low-power devices.
At the same time, keypoint-based object detection methods have gained significant popularity in recent papers as
they are simpler, faster, and more accurate than the corresponding anchor-based detectors.

Previous approaches, such as~\cite{liu2016ssd, redmon2016you}, required anchor boxes
to be manually designed in order to train detectors.
A series of anchor-free object detectors were then developed, aiming to predict the bounding box
keypoints, rather than trying to fit an object to an anchor.
Law and Deng proposed CornerNet~\cite{law2018cornernet}, a novel anchor-free framework which detected objects like a pair of corners,
and predicted class heatmaps, pair embeddings, and corner offsets on each position of the feature map.
Class heatmaps calculated the probability of a corner, corner offsets were used
to regress the corner location, and the pair embeddings grouped a pair of corners that belong to the same object.
Without relying on manually designed anchors to match objects, CornerNet achieved
a significant performance improvement on the MS COCO datasets.
Subsequently, several other variants of keypoint detection-based one-stage detectors have been developed~\cite{Tian_2019_ICCV, Zhou_2019_CVPR}.

In~\cite{DeepMark}, authors naturally treat landmark estimation as a pose estimation
problem, and use keypoint-based multi-target architecture CenterNet~\cite{CenterNet}
with slight modifications to handle both object detection and landmark estimation.
We further develop that idea in this paper, and propose several improvements to accelerate
the training process and enhance performance.

Despite state-of-the-art performance in object detection and keypoint estimation tasks, CenterNet's approach
can produce overconfident incorrect predictions, leading to a decrease in performance metrics.
A separate line of research exists in the field of deep learning which aims to solve the problem of estimating the uncertainty of model predictions.
In this regard, a number of papers show how the use of various techniques, aimed at obtaining a more accurate estimate of the prediction uncertainty,
can improve the quality of detection~\cite{miller2017dropout, Kraus_2019, TruongLe2018UncertaintyEF}.
The most popular method for estimating model uncertainty, MC Dropout, requires several forward passes through a network to be performed,
significantly increasing the inference time.
In~\cite{TruongLe2018UncertaintyEF}, a sample-free method for uncertainty estimation in anchor-based detectors was presented
which uses a set of observations that relate to the same object, according to Jaccard overlap (before applying NMS).
Motivated by this idea, we created a way to improve CenterNet's performance by
considering a set of predictions related to a single object instead of point estimation.

\section{Proposed approach}\label{sec:proposed-approach}
\begin{figure}[t]
	\centering
	\includegraphics[width=\columnwidth]{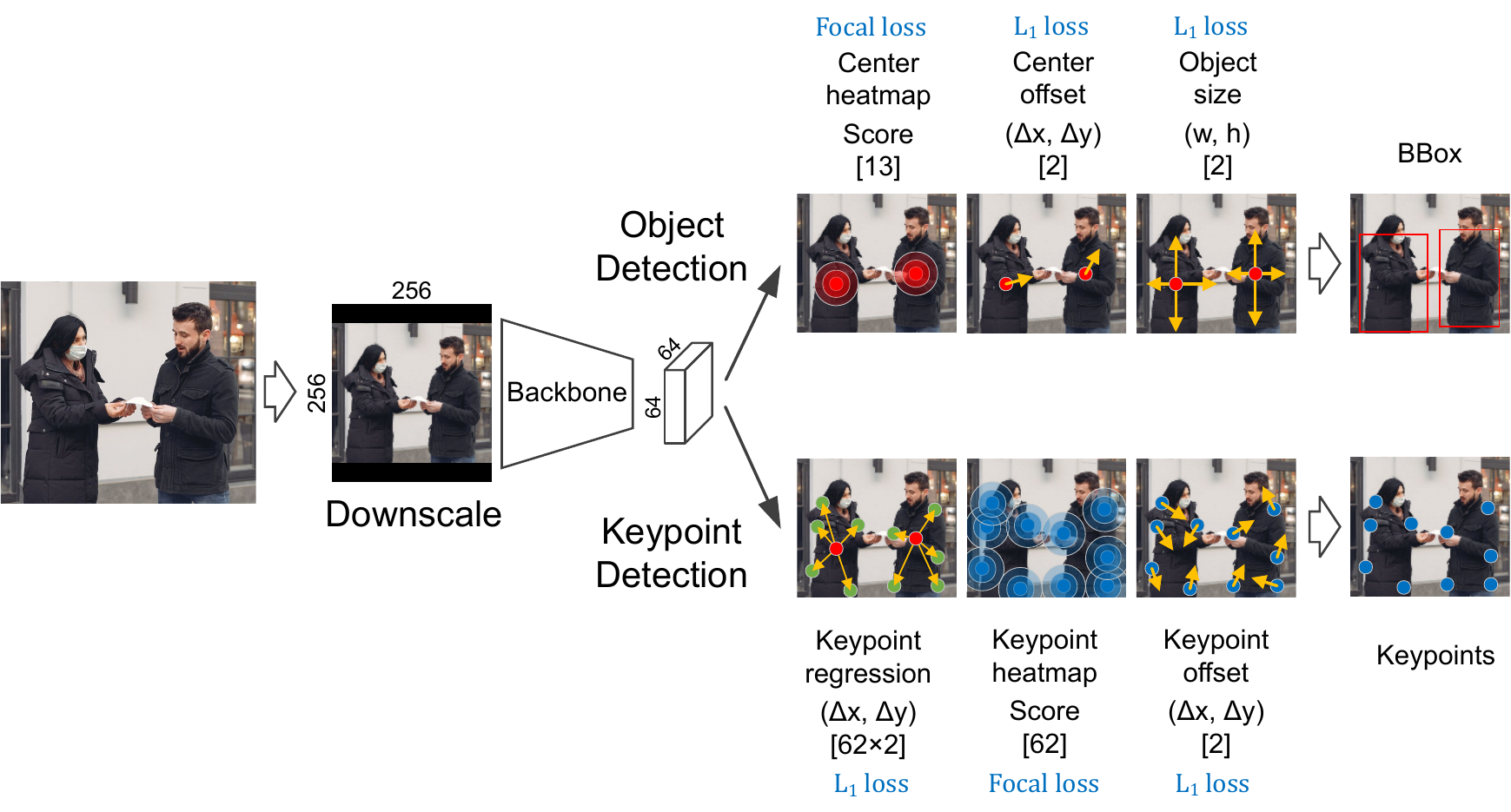}
	\caption{Scheme of the proposed approach.}
	\label{fig:scheme}
\end{figure}

Our approach is based on the CenterNet~\cite{CenterNet} architecture, and
simultaneously solves the following two tasks: object detection and keypoint location estimation.
The architecture is presented in Figure~\ref{fig:scheme}.
A backbone downsamples the input image by a factor of 4 to produce a feature map,
which is then processed to obtain object bounding boxes and corresponding keypoints.

As the DeepFashion2 dataset contains 13 classes,
13 channels are used in the center heatmap to predict the probabilities for each pixel being the object center for each class.
The center of an object is defined as the center of a bounding box,
and a ground truth heatmap is generated by applying a Gaussian function at each object's center.
Two additional channels in the output feature map -- $\bigtriangleup$x and $\bigtriangleup$y -- are used to refine the center coordinates,
and both width and height are predicted directly.

Another branch handles the fashion landmark estimation task,
which involves estimating 2D keypoint locations for each item of clothing in one image.
The coarse locations of the keypoints are regressed as relative displacements from the box center (keypoint regression in Figure~\ref{fig:scheme}).
Consequently, if a certain pixel has already been classified as an object center,
one can take the values in the same spatial location from this tensor and interpret them as vectors to keypoints.
The keypoint position obtained through regression is not entirely accurate, so an additional heatmap with
probabilities is used for each keypoint type to refine a keypoint location.
Here, a local maximum with high confidence in the heatmap is used as a refined keypoint position.
Similar to the detection branch, two additional channels -- $\bigtriangleup$x and $\bigtriangleup$y -- are used to obtain more precise landmark coordinates.
During model inference, each coarse keypoint location is replaced with the closest refined keypoint position.
As such, keypoints that belong to the same object are grouped together.

\begin{figure}[t]
	\centering
	\includegraphics[width=\columnwidth]{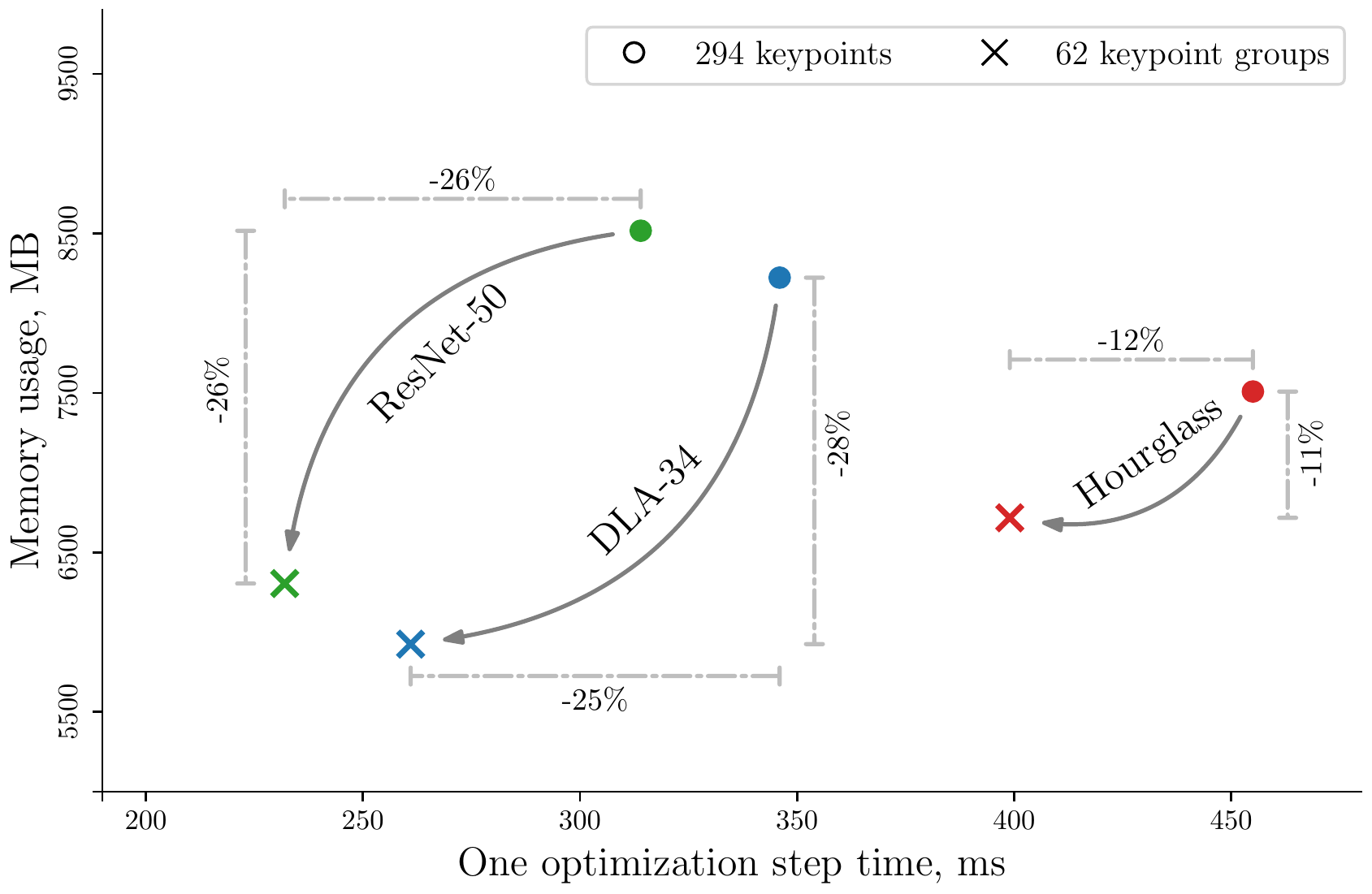}
	\caption{GPU memory consumption and training iteration time using an RTX 2080ti and PyTorch 1.4.
	The input resolution is $\medmuskip=0mu 256\times256$, while the batch size is 32 for both DLA-34 and ResNet-50, and 8 for Hourglass.
	The time in ms was measured for 1 optimization step: batch loading to GPU, forward pass and backward pass.
	GPU memory was measured using the nvidia-smi tool.}
	\label{fig:time-mem-plot}
\end{figure}

\subsection{Semantic keypoint grouping}\label{subsec:semantic-keypoint-grouping}

One of the first steps involved in solving keypoint detection tasks is defining the model output.
The number of keypoints for every category varies from 8 for a skirt, to 39 for a long sleeve outerwear in the DeepFashion2 dataset,
and the total number of unique keypoints is 294.
The straightforward approach is to concatenate keypoints from every category and deal with each one separately.
Directly predicting 294 keypoints leads to a huge number of output channels:
$901 = 13 + 2 + 2 + 294 \cdot 2 + 294 + 2$ ($13$ -- center heatmap, $2$ -- center offset, $2$ -- object size,
$294 \cdot 2$ -- keypoint regression, $294$ -- keypoint heatmap, $2$ -- keypoint offset).

It is evident that certain clothing landmarks are actually a subset of others.
For example, shorts do not require unique keypoints as they can be represented by a subset of trouser keypoints.
The semantic grouping rule is defined as follows (Figure~\ref{fig:keypoints}):
identical semantic meaning keypoints (such as collar center and top sleeve edge) with different categories can be merged into one group.
This approach enables the formation of 62 groups and reduces the number of output channels from 901 to
$205 = 13 + 2 + 2 + 62 \cdot 2 + 62 + 2$ (see Figure~\ref{fig:scheme} for details).

The semantic grouping approach reduces training time and memory consumption by up to 26\% and 28\%, respectively,
without notable decrease in accuracy (see Figure~\ref{fig:time-mem-plot} and \autoref{tab:grouping-accuracy}).
In addition, the latter reduction enables the use of larger batches during model training.

The semantic grouping approach also reduces inference time and memory consumption by up to 26\% and 37\%,
respectively (see section~\ref{subsec:mobile-phone} for details).
An in-depth analysis of the grouping approach is presented in~\cite{EfficientGrouping}.

\begin{table}
    \begin{center}
		\setlength\tabcolsep{6.0pt} 
        \begin{tabular}{|l|c|c|c|c|}
            \hline
			\multirow{2}{*}{Backbone} & \multicolumn{2}{c|}{$\mathit{mAP}_{pt}$} & \multicolumn{2}{c|}{$\mathit{mAP}_{box}$} \\\cline{2-5}
			                          & 294 kps & 62 kps & 294 kps & 62 kps \\
			\hline
			ResNet-50                 & 0.377   & 0.365  & 0.629   & 0.627 \\
			\hline
			mDLA-34                   & 0.434   & 0.422  & 0.649   & 0.651 \\
			\hline
			DLA-34                    & 0.466   & 0.456  & 0.679   & 0.679 \\
			\hline
        \end{tabular}
    \end{center}
	\caption{The detection accuracy for different backbones with (62~kps) and without (294~kps) semantic keypoint grouping.}
	\label{tab:grouping-accuracy}
\end{table}

\subsection{Post-processing techniques}\label{subsec:post-processing-techniques}

We have developed 4 post-processing techniques that increase the model's accuracy without compromising performance.

\subsubsection{Center rescoring}

\begin{figure}[t]
	\centering
	\includegraphics[width=0.9\columnwidth]{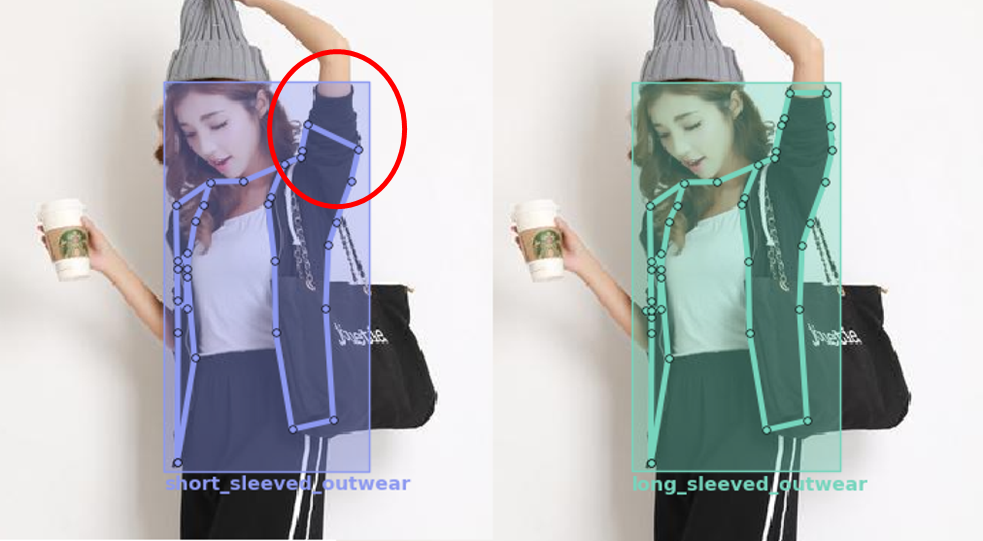}
	\caption{Detection results for two different classes.
	The keypoint score for the right class has a higher value.
	Short sleeved outwear on the left has $Score_{bbox}=0.31$, $Score_{kps}=0.28$.
	Long sleeved outwear on the right has $Score_{bbox}=0.29$, $Score_{kps}=0.36$.
	}
	\label{fig:technique_1}
\end{figure}

It is evident that keypoint scores for the wrong category may have low confidence (see Figure~\ref{fig:technique_1}).

We use this insight to recalculate the detection confidence score using keypoint scores from the keypoint heatmap.
Let $Score_{bbox}$ be the original detection confidence score from the center heatmap, and $Score_{kps}$ be the average score
of the refined keypoints for the predicted category from the keypoint heatmap.
The final detection confidence scores are calculated as a linear combination of the bounding box score and the average keypoint score:
\begin{equation}
Score = \alpha \cdot Score_{bbox} + (1-\alpha) \cdot Score_{kps}, \label{eq:1}
\end{equation}
where $\alpha \in [0,1]$.

\subsubsection{Heatmap rescoring with Gaussian kernel}

\begin{figure}[t]
	\centering
	\includegraphics[width=\columnwidth]{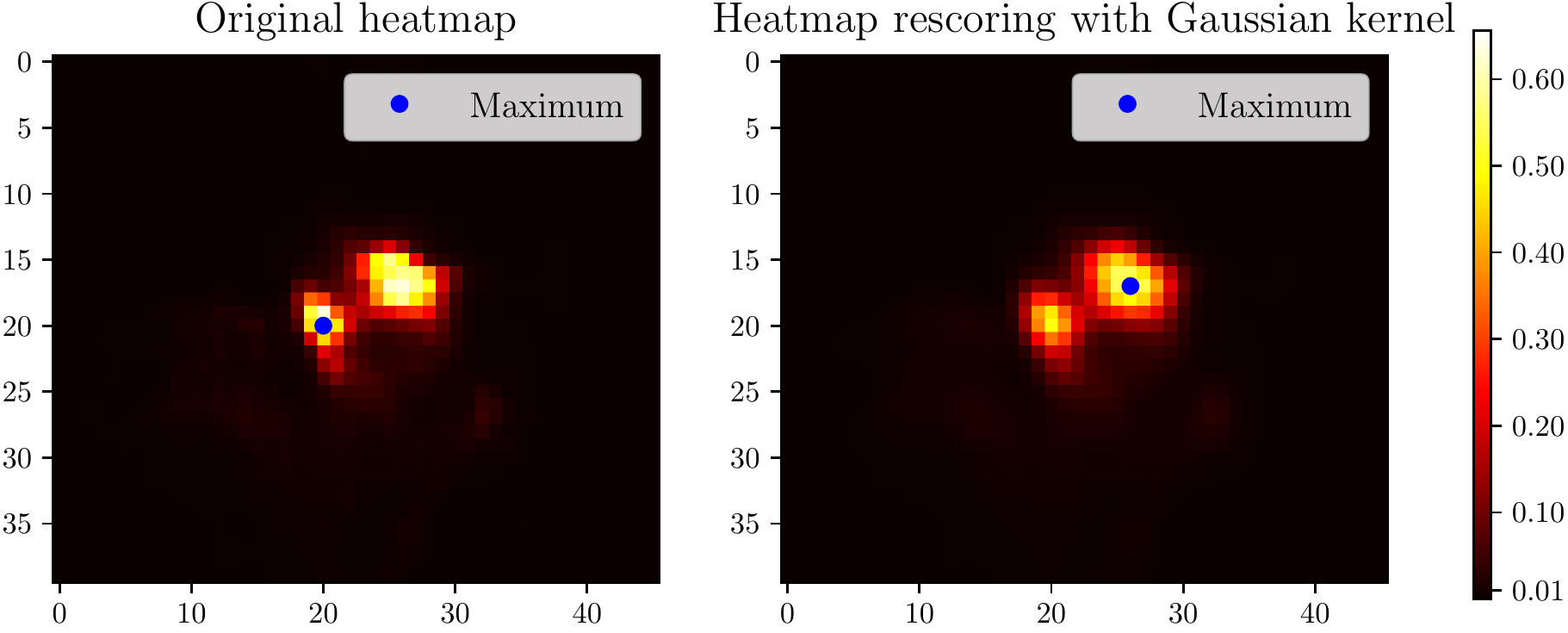}
	\caption{A heatmap with center scores and a heatmap after rescoring with Gaussian kernel ($\sigma=0.45$).
	}
	\label{fig:technique_2}
\end{figure}

Applying smoothing operators on heatmaps is quite a common technique.
For instance, it is used in CenterNet to generate ground truth labels for a heatmap during the training stage.
We propose a general approach that can be applied to any keypoint-based architecture during the inference stage as well.

Let $H$ be a heatmap with center or keypoint scores.
Taking the training procedure into account, you can expect the 8-connected neighbors of each item to be related to the same object,
and this fact can be used to improve the estimation of each heatmap value (see Figure~\ref{fig:technique_2}).
Consequently, we have applied the following formula:
\begin{equation}
\hat{H} = H \circledast G(\sigma), \label{eq:2}
\end{equation}
where $\circledast$ is the convolution operation, and ${G(\sigma)}$ is the ${3\times3}$ Gaussian kernel with the standard deviation $\sigma$.
Experimental results show that in our model, the proposed technique improves the localization of peaks and their values
that correspond to object centers or keypoints and their scores.
A similar operation has been considered in~\cite{DARK} as a part of the proposed method.

\subsubsection{Keypoint location refinement}

\begin{figure}[t]
	\centering
	\includegraphics[width=\columnwidth]{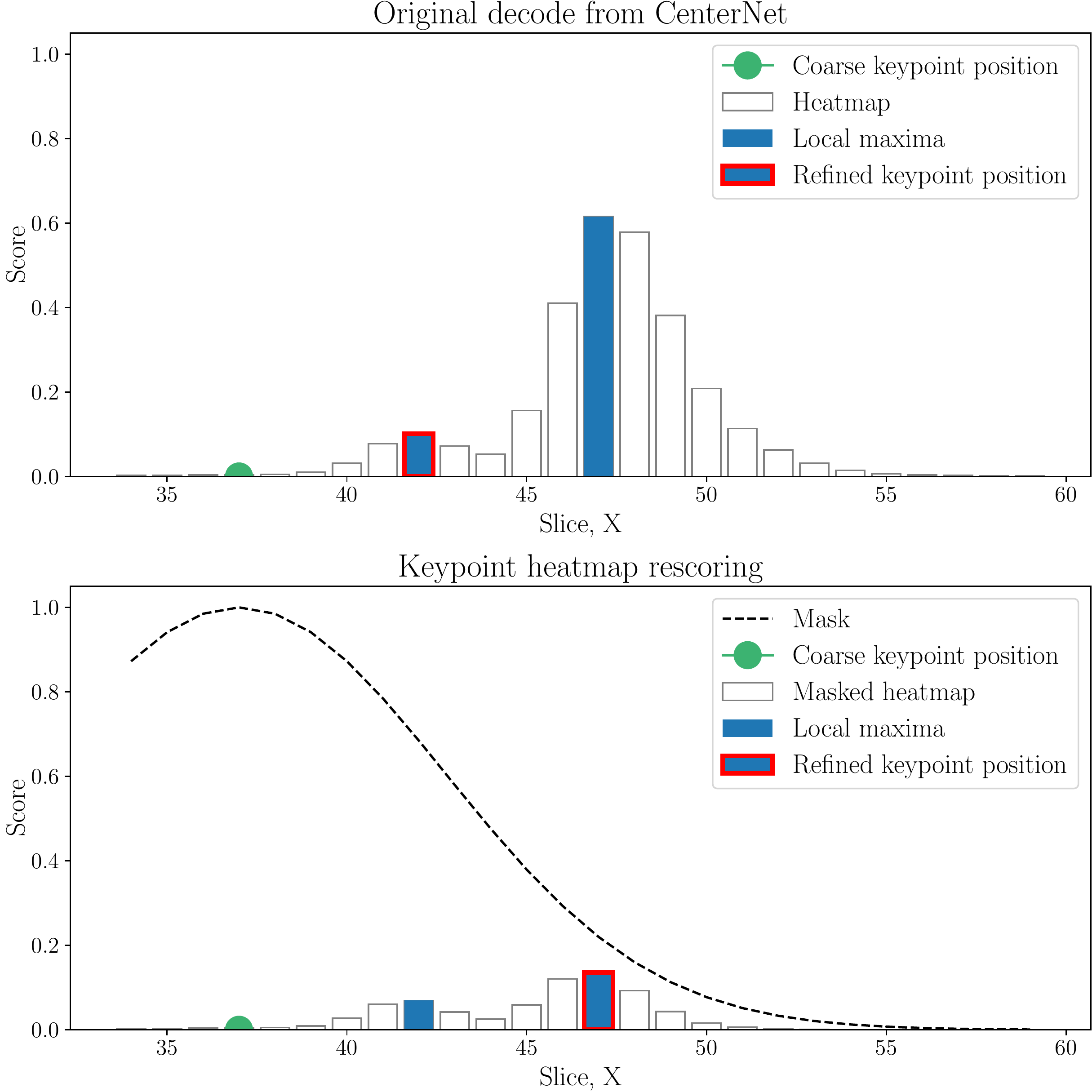}
	\caption{
	In CenterNet,
      the refined location is defined as the closest point to the coarse keypoint position,
	which in turn is defined as a local maximum on the heatmap.
     This can lead to errors when the closest prediction is not actually the best one (as evident in the diagram above).
     In the proposed technique, the refined location is determined as a global maximum of the keypoint heatmap rescored by the Gaussian mask,
	improving keypoint localization.
	}
	\label{fig:technique_4}
\end{figure}

\begin{figure}[t]
	\centering
	\includegraphics[width=0.32\columnwidth]{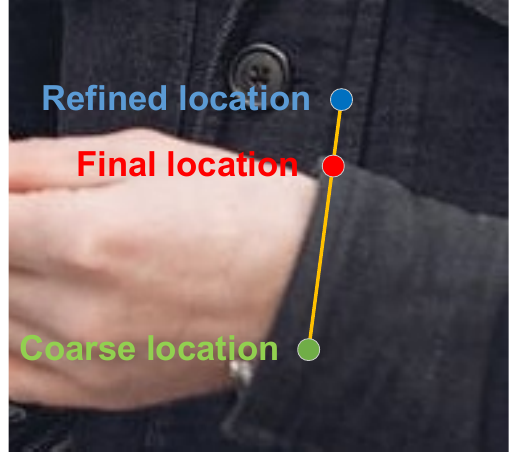}
	\caption{A linear combination of the refined and coarse keypoint positions.
	}
	\label{fig:technique_3}
\end{figure}

The third technique further corrects keypoint locations by calculating a linear combination of the refined and
coarse keypoint positions (see Figure~\ref{fig:technique_3}).
Our experiments have shown a small improvement when using this technique, with no impact on performance.

Let $(x, y)_{refined}$ be the refined keypoint location from the heatmap and $(x, y)_{coarse}$ be coarse positions predicted as offsets from object centers.
The final keypoint locations are calculated through the following expression:
\begin{equation}
(x, y) = \gamma \cdot (x, y)_{refined} + (1-\gamma) \cdot (x, y)_{coarse}, \label{eq:3}
\end{equation}
where $\gamma \in [0,1]$.

\subsubsection{Keypoint heatmap rescoring}

In the fourth technique we propose the addition of a penalty to the keypoint score in proportion to the distance from a coarse keypoint position with the Gaussian function.
Let $mask$ be a heatmap with zero values by default.
We set $1$ into the $mask$ in the coarse keypoint position and fill neighbor values with 2D Gaussian function with standard deviation ${sigma=\min(width, height)/9}$,
where $width$ and $height$ are the object size (see the second image in~Figure~\ref{fig:technique_4}).

The keypoint heatmap is rescored through the following expression:
\begin{equation}
\hat{H}_{kps} = H_{kps} \cdot mask. \label{eq:4}
\end{equation}


\section{Results}\label{sec:results}
All experiments were performed on the publicly available DeepFashion2 Challenge
dataset~\cite{DeepFashion2}, which contains 191,961 images in the training set
and 32,153 images in the validation set.

\begin{figure}[t]
	\centering
	\includegraphics[width=\columnwidth]{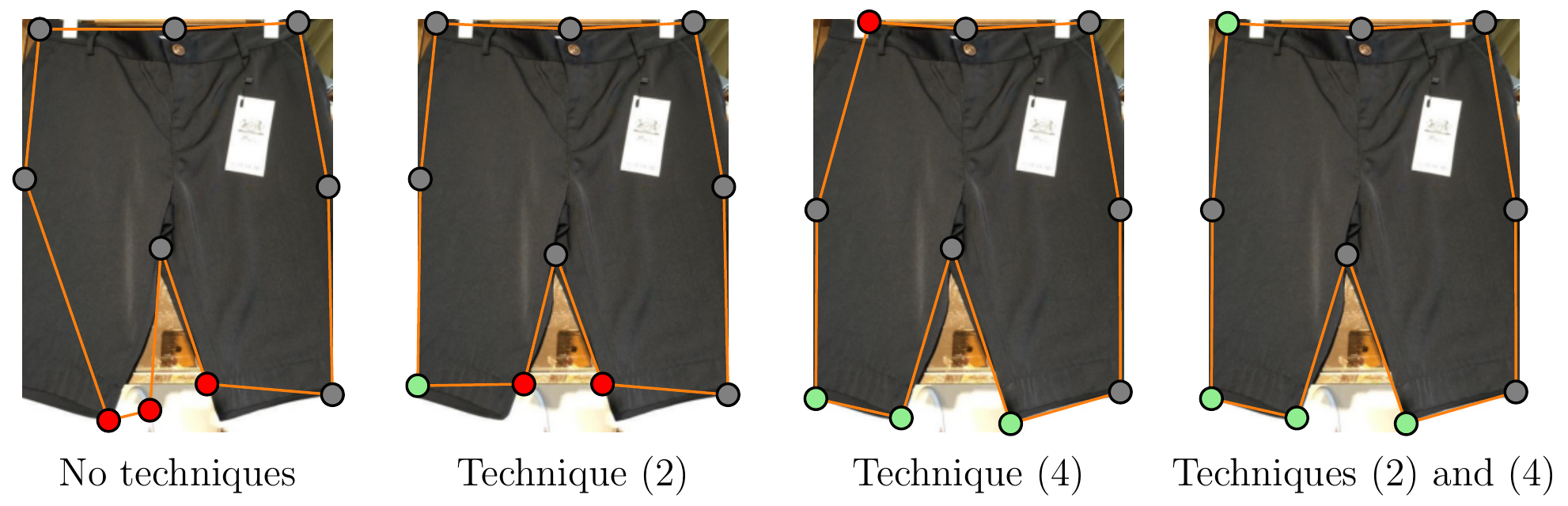}
	\caption{
		The impact of post-processing techniques on keypoint detection accuracy.
		Red dots highlight incorrect detections, and green dots highlight fixes by post-processing techniques.
	}
	\label{fig:shorts}
\end{figure}

\subsection{Post-processing techniques}\label{subsec:results-post-processing-techniques}

We considered 5 fast post-processing techniques: bounding box non-maximum suppression and 4 techniques from section~\ref{subsec:post-processing-techniques}.
The individual and combined effectiveness of each technique is shown in \autoref{tab:single-experiments} and Figure~\ref{fig:shorts}.

Certain techniques can increase $\mathit{mAP}_{pt}$ and reduce $\mathit{mAP}_{box}$ simultaneously.
Note that bounding box detection and keypoint estimation results for the same object may have different IoU and OKS with the ground truth, for example, when bounding box was detected correctly, but keypoints were not.
In this case, technique~\eqref{eq:1} involves lowering a score for false positive keypoints, which is advisable.
The corresponding true positive bounding box also suffers from this lowered score.

\begin{figure}[t]
	\centering
	\includegraphics[width=\columnwidth]{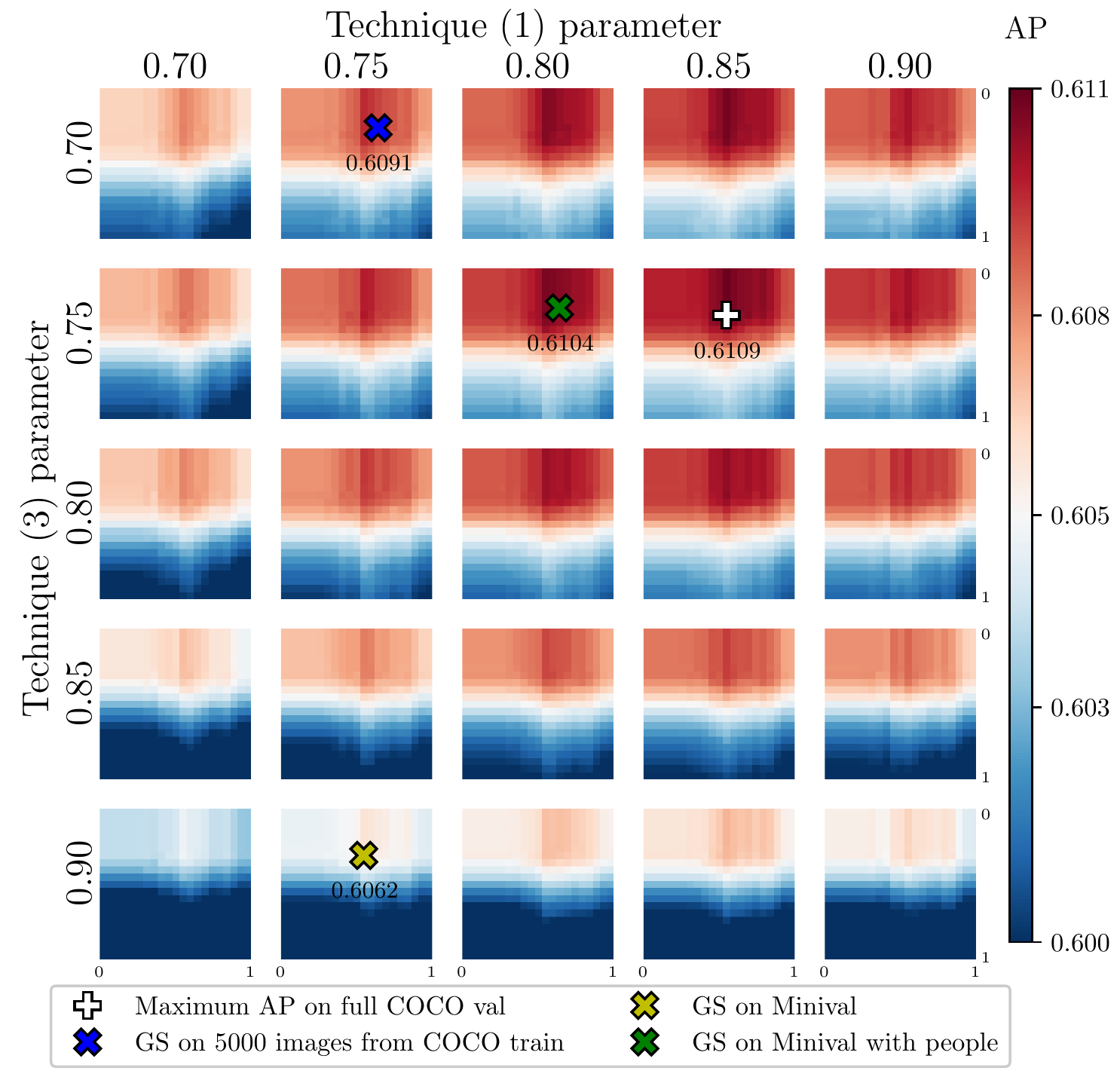}
	\caption{
    AP on the COCO validation dataset with different post-processing technique parameters.
	Every heatmap shows AP for technique~\eqref{eq:2} with step 0.05 in a range from 0 to 1
		(center heatmap ${\sigma}$ goes on the X-axis and keypoint heatmap ${\sigma}$ goes on the Y-axis).
	Optimal values found with grid search (GS) on different datasets are shown.
	}
	\label{fig:params_coco_zoom}
\end{figure}

We also evaluated proposed techniques on the COCO keypoint detection task.
Here, we took the DLA-34 model for human keypoint detection on the COCO dataset with 0.589 AP from~\cite{CenterNet}, and applied 4 techniques from section~\ref{subsec:post-processing-techniques}.
Three datasets were selected for hyperparameter optimization:
5000 random images from the COCO train dataset;
a subset of 100 images from the COCO validation dataset, of which only 61 images contain people (Minival);
a subset of 100 images with people from the COCO validation dataset (Minival with people).

Each post-processing technique parameter was determined through grid searching with step 0.05 within the range 0 to 1,
resulting in 21 experiments for a single parameter.
Grid search on a representative subset enables accuracy very close to the global maximum (see Figure~\ref{fig:params_coco_zoom}),
and the accuracy was increased by 0.021 AP (from 0.589 AP to 0.6104 AP) without model retraining.
On Minival with people we determined the following optimal parameters: ${\alpha = 0.75}$, center heatmap ${\sigma = 0.25}$, keypoint heatmap ${\sigma = 0.65}$, ${\gamma = 0.8}$.

\subsection{DeepFashion2 Challenge}\label{subsec:results-deepFashion2-challenge}

We used the CenterNet MS COCO model for object detection as the initial checkpoint and performed
experiments with the Hourglass backbone and Adam the optimizer to achieve the best results (\autoref{tab:compare-approaches}) for
object detection and keypoint estimation tasks on the DeepFashion2 validation dataset.
The Hourglass~$\medmuskip=0mu 512\times512$ model was trained for 100 epochs with a batch size of 46 images.
The learning rate schedule is: 1e-3 - 65 epochs, 4e-4 - 20 epochs, 4e-5 - 15 epochs.
The Hourglass~$\medmuskip=0mu 768\times768$ model was fine-tuned from Hourglass~$\medmuskip=0mu 512\times512$ for 25 epochs, with a batch size of 22 images: 2e-5 - 20 epochs, 1e-5 - 5 epochs.

\begin{table}
    \begin{center}
		\setlength\tabcolsep{3.0pt} 
        \begin{tabular}{|l|c|c|c|}
            \hline
			\multirow{2}{*}{Post-processing} & \multirow{2}{*}{$\mathit{mAP}_{pt}$} & \multirow{2}{*}{$\mathit{mAP}_{box}$} & Total time,\\
			                                 &                                      &                                       &  ms \\
            \hline
            Baseline                  & 0.422 & 0.651 &  54.50 \\
			\hline
			NMS                       & 0.422 & 0.652 &  54.57 \\
			\hline
			Technique~\eqref{eq:1}    & 0.428 & 0.650 &  54.61 \\
			\hline
			Technique~\eqref{eq:2}    & 0.430 & 0.651 &  54.44 \\
			\hline
			Technique~\eqref{eq:3}    & 0.431 & 0.651 &  54.81 \\
			\hline
			Technique~\eqref{eq:4}    & 0.430 & 0.651 &  55.95 \\
            \hline
			\hline
			Combined    & 0.445 & 0.652 &  56.76 \\
            \hline
        \end{tabular}
    \end{center}
	\caption{Different post-processing techniques applied independently to mDLA-34 $\medmuskip=0mu 256\times256$ on the DeepFashion2 validation dataset.
	The technique numbers correspond to the numbers in section~\ref{subsec:post-processing-techniques}.
	Total processing time is measured on a Huawei P40 Pro.
	Technique~\eqref{eq:2} works faster than the base version, as it significantly reduces the number of local maxima and increases the speed of further decoding.
	The last line contains metrics for combined techniques.
	}
	\label{tab:single-experiments}
\end{table}

\begin{table}
	\begin{center}
		\begin{tabular}{|c|c|c|}
			\hline
			Approach                       & $\mathit{mAP}_{pt}$ & $\mathit{mAP}_{box}$ \\
			\hline
			Mask R-CNN~\cite{DeepFashion2} & 0.529               & 0.638 \\
			DeepMark~\cite{DeepMark}       & 0.532               & 0.723 \\
			DAFE~\cite{Chen_2019_ICCV}     & 0.549               & ---  \\
			Aggregation and Finetuning~\cite{Lin2020AggregationAF}     & 0.614               & 0.764  \\
			\hline
			Hourglass $\medmuskip=0mu 512\times512$      & 0.583               &  0.735 \\
			Hourglass $\medmuskip=0mu 768\times768$      & 0.591            &  0.737 \\
			\hline
		\end{tabular}
	\end{center}
	\caption{Accuracy comparison between the proposed and alternative approaches with the DeepFashion2 validation dataset.
	All models are trained with the released DeepFashion2 Challenge dataset.}
	\label{tab:compare-approaches}
\end{table}

To optimize post-processing technique parameters with grid search (subsection~\ref{subsec:results-post-processing-techniques})
we defined a small validation subset (1285 images) from the DeepFashion2 validation dataset.
The following parameters were used for the Hourglass $\medmuskip=0mu 768\times768$: ${\alpha = 0.8}$, center heatmap ${\sigma = 0.45}$,
keypoint heatmap ${\sigma = 0.90}$, ${\gamma = 0.75}$.

With the Hourglass $\medmuskip=0mu 768\times768$ model we achieved the second place in the DeepFashion2 Challenge 2020 with 0.582 $\mathit{mAP}$ on the test dataset.

During all experiments, our target was to increase the keypoint estimation accuracy instead of the object detection accuracy.
Consequently, object detection increased by only 0.042 $\mathit{mAP}_{box}$, but all the techniques added more than 0.07 to $\mathit{mAP}_{pt}$ (see~\autoref{tab:all-experiments}).

\subsubsection{Multi-inference strategies}\label{subsec:multi-inference-strategies}

We have considered 2 extra inference strategies: fusing model outputs from original and flipped images with equal weights;
and fusing model results with the original image downscaled/upscaled through certain multipliers.
While these proposed techniques increase accuracy, they require several model inferences, and this significantly impacts processing time.

\begin{figure}[t]
	\centering
	\includegraphics[width=\columnwidth]{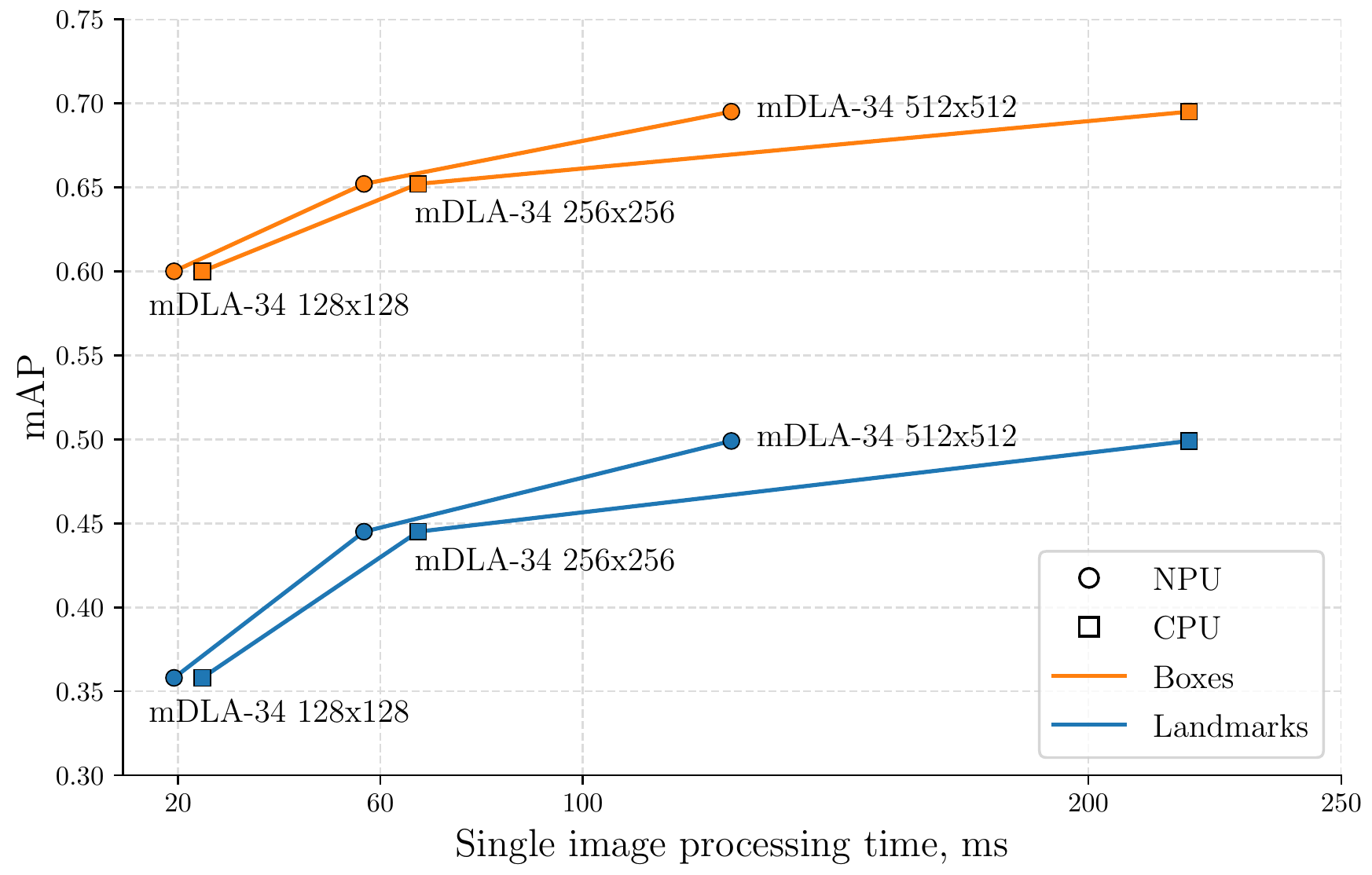}
	\caption{
		Speed/accuracy trade-off for object detection and landmark estimation on the DeepFashion2 validation dataset.
		Total processing time (pre-procesing + inference + post-processing) in ms was measured on a Huawei P40 Pro with MNN on CPU and with HiAI DDK on NPU.
	}
	\label{fig:plot_phone}
\end{figure}

\begin{table}
    \begin{center}
		\setlength\tabcolsep{1.0pt} 
        \begin{tabular}{|l|c|c|c|c|}
            \hline
			Metric & \multicolumn{2}{c|}{$\mathit{mAP}_{pt}$} & \multicolumn{2}{c|}{$\mathit{mAP}_{box}$} \\\cline{2-5}
			\hline
			Resolution & $\medmuskip=0mu 512\times512$ & $\medmuskip=0mu 768\times768$ & $\medmuskip=0mu 512\times512$  & $\medmuskip=0mu 768\times768$  \\
            \hline
            Baseline             & 0.529 & 0.520 & 0.720 & 0.695 \\
			\hline
			$+$ Post-processing & 0.545 & 0.540 & 0.713 & 0.698 \\
			\hline
            $+$ NMS          & 0.548 & 0.549 & 0.718 & 0.712 \\
			\hline
            $+$ Flip         & 0.561 & 0.563 & 0.731 & 0.727 \\
			\hline
            $+$ Multiscale   & 0.568 & 0.578 & \textbf{0.735} & \textbf{0.737} \\
            \hline
			$+$ PoseFix      & \textbf{0.583} & \textbf{0.591} & \textbf{0.735} & \textbf{0.737} \\
            \hline
        \end{tabular}
    \end{center}
	\caption{Clothing detection and landmark estimation.
	Hourglass with $\medmuskip=0mu 512\times512$ and $\medmuskip=0mu 768\times768$ resolution were used in the experiments.
	The next technique is added to each of the previous ones,
	and the post-processing refers to the application of all techniques from section~\ref{subsec:post-processing-techniques}.
	We applied the following multipliers for the multiscale technique: 0.85, 0.95, 1.1.
	Our target was to increase the keypoint estimation accuracy (also consider the note from~\ref{subsec:results-post-processing-techniques}),
	so $\mathit{mAP}_{box}$ accuracy is dropped after applying some of our techinques.}
	\label{tab:all-experiments}
\end{table}

\subsubsection{Keypoint Refinement Network}\label{subsec:keypoint-refinement-network}

At the final stage, detection results are refined with the PoseFix~\cite{moon2018posefix} model-agnostic pose refinement method,
which learns the typical error distributions of any other pose estimation method and corrects mistakes at the testing stage.

We trained a set of 13 PoseFix models using the number of classes in the DeepFashion2 dataset,
and the inference results of our method on the training set are used to train each of the 13 models.
Subsequently, we applied the trained PoseFix models to the result.

\subsection{Experiments on a mobile phone}\label{subsec:mobile-phone}

The DLA-34 model contains DCN~\cite{DCN} layers by default.
This increases model accuracy by 2-3 mAP, but most inference frameworks do not currently support DCN.
In order to run the DLA-34 model on a mobile phone, we replaced all DCN layers with conventional convolution layers $3\times3$.
This model is referred to as mDLA-34, and it contains 19.5 millions parameters.
The number of floating point operations for the model is presented in~\autoref{tab:flops}.

\begin{figure}[t]
	\centering
	\includegraphics[width=\columnwidth]{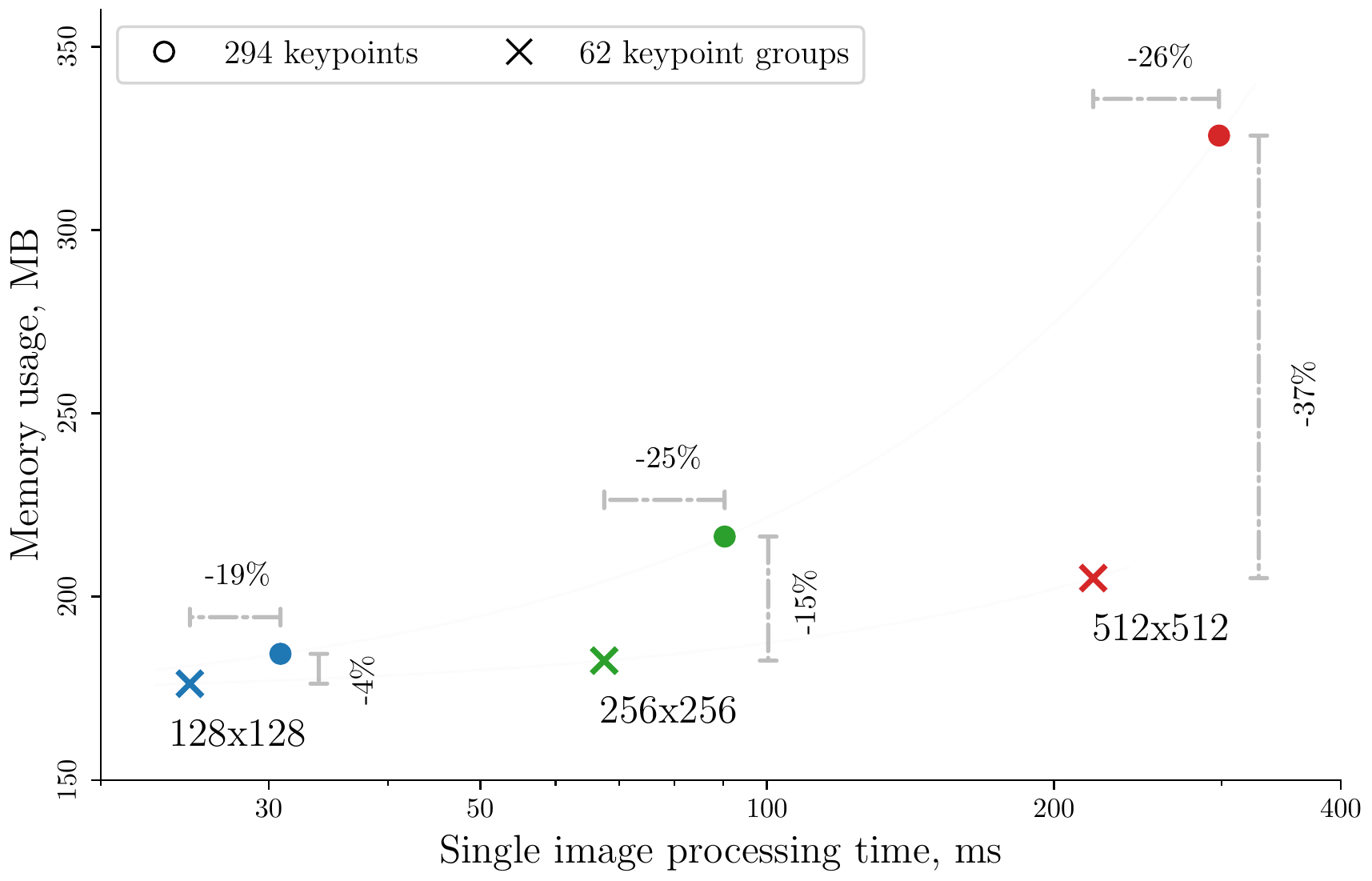}
	\caption{
    Single image processing time and memory consumption with mDLA-34 on a Huawei P40 Pro with MNN.
	Total processing time (pre-procesing + inference + post-processing) and memory usage during inference are shown
	for models with direct prediction of 294 keypoints and a prediction of 62 keypoint groups.
	}
	\label{fig:time_mem_phone_plot}
\end{figure}

The Huawei P40 Pro smartphone (complete with Kirin 990) is used for our experiments.
To run the model on CPU we use MNN~\cite{alibaba2020mnn}~v1.0.2 (number of threads = 4), and on NPU we use HiAI~DDK~v100.310.011.
Both NPU and CPU can run mDLA-34 at 14 FPS with relatively high accuracy (see Figure~\ref{fig:plot_phone}).

\begin{figure*}
	\centering
	\includegraphics[width=\textwidth]{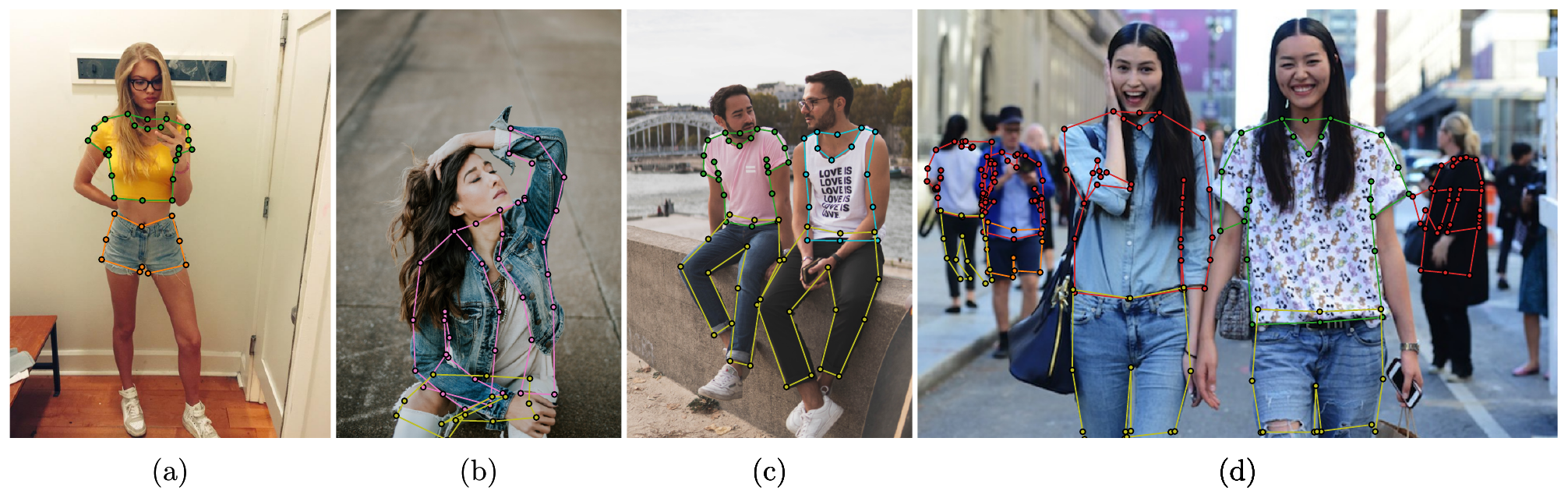}
	\caption{
		Examples of fashion landmark detection with mDLA-34~$\medmuskip=0mu 256\times256$ with all post-processing techniques.
		A simple pose without any occlusion is not a major challenge for even the $\medmuskip=0mu 256\times256$ model.
		Keypoints are detected quite accurately for images featuring a single person (a) or multiple people (c).
		However, uncommon hand positions (b) can lead to errors in the elbow area.
		As the subject's legs are not fully visible in image (b), this leads to both erroneous determination of the trouser type
		and errors in the keypoint positions.
		In addition, clothing detection on distant or blurred subjects (d) works poorly in some cases.
	}
	\label{fig:cases}
\end{figure*}

\begin{table*}
    \begin{center}
		\setlength\tabcolsep{2.0pt} 
        \begin{tabular}{|l|c|c|c|c|c|c|c|c|c|c|c|c|c|}
            \hline
			& \multicolumn{3}{c|}{Scale} & \multicolumn{3}{c|}{Occlusion} & \multicolumn{3}{c|}{Zoom-in} & \multicolumn{3}{c|}{Viewpoint} & \multirow{2}{*}{Overall} \\\cline{2-13}
			                                      & Small & Moderate & Large & Slight & Medium & Heavy & No    & Medium & Large & No wear & Frontal & Side or back & \\
			\hline
            \multirow{2}{*}{$\mathit{mAP}_{pt}$}  & 0.408 & 0.560    & 0.467 & 0.509  & 0.502  & 0.209 & 0.536 & 0.440  & 0.315 & 0.431   & 0.550   & 0.409        & 0.507 \\
			                                      & 0.346 & 0.497    & 0.434 & 0.490  & 0.421  & 0.132 & 0.473 & 0.393  & 0.295 & 0.418   & 0.482   & 0.345        & 0.445 \\
			\hline
			$\mathit{mAP}_{box}$                  & 0.574 & 0.683    & 0.667 & 0.691  & 0.636  & 0.285 & 0.663 & 0.641  & 0.592 & 0.669   & 0.669   & 0.614        & 0.652 \\
            \hline
        \end{tabular}
    \end{center}
	\caption{The results for landmark estimation with mDLA-34~$\medmuskip=0mu 256\times256$ with all post-processing techniques
	on different validation subsets, including scale, occlusion, zoom-in, and viewpoint.
	Results of evaluation on visible landmarks only and evaluation on both visible and
	occluded landmarks are separately shown in each row for $\mathit{mAP}_{pt}$.}
	\label{tab:mdla-details}
\end{table*}

We also estimated the influence on the keypoint grouping approach to memory consumption and execution time (see Figure~\ref{fig:time_mem_phone_plot}).
Here, the grouping approach only impacts the last part of a neural network and a post-processing step.
The higher the input resolution, the greater the performance and memory gains.
The keypoint grouping approach works 25\% faster and requires 15\% less memory with the mDLA-34~$256\times256$ model on a Huawei P40 Pro, compared to the same model without keypoint grouping.

mDLA-34~$256\times256$ performance on the different validation subsets and for each class is presented in~\autoref{tab:mdla-details} and \autoref{tab:mdla-by-class-accuracy}.
Figure~\ref{fig:cases} demonstrates examples of the model performance on several use-cases.

\begin{table}
	\begin{center}
		\setlength\tabcolsep{3.0pt} 
		\begin{tabular}{|c|c|c|c|c|}
			\hline
			\multirow{2}{*}{Resolution} & Model                       & Post-processing & \multirow{2}{*}{Decode}  & \multirow{2}{*}{Total} \\
			                            & inference                   & techniques      &                          & \\
			\hline
			$128\times128$              & \multicolumn{1}{r|}{2.423}  & 0.0008          & 0.0004                   &  \multicolumn{1}{r|}{2.424} \\
			\hline
			$256\times256$              & \multicolumn{1}{r|}{9.691}  & 0.0028          & 0.0015                   &  \multicolumn{1}{r|}{9.696} \\
			\hline
			$512\times512$              & \multicolumn{1}{r|}{38.765} & 0.0113          & 0.0058                   & \multicolumn{1}{r|}{38.783} \\
			\hline
		\end{tabular}
	\end{center}
	\caption{The number of floating point operations in GFLOPs for mDLA-34 inference, post-processing techniques, and decode.}
	\label{tab:flops}
\end{table}

\begin{table}[h]
    \begin{center}
		\setlength\tabcolsep{6.0pt} 
        \begin{tabular}{|l|c|c|}
            \hline
			Class & $\mathit{AP}_{pt}$ & $\mathit{AP}_{box}$ \\
			\hline
			Short sleeve top     & 0.606 & 0.804 \\
			\hline
			Long sleeve top      & 0.507 & 0.724 \\
			\hline
			Short sleeve outwear & 0.175 & 0.347 \\
			\hline
			Long sleeve outwear  & 0.458 & 0.724 \\
			\hline
			Vest                 & 0.426 & 0.679 \\
			\hline
			Sling                & 0.297 & 0.422 \\
			\hline
			Shorts               & 0.545 & 0.721 \\
			\hline
			Trousers             & 0.443 & 0.739 \\
			\hline
			Skirt                & 0.481 & 0.740 \\
			\hline
			Short sleeve dress   & 0.586 & 0.721 \\
			\hline
			Long sleeve dress    & 0.375 & 0.542 \\
			\hline
			Vest dress           & 0.481 & 0.710 \\
			\hline
			Sling dress          & 0.400 & 0.605 \\
			\hline
        \end{tabular}
    \end{center}
	\caption{Accuracy by class for mDLA-34~$\medmuskip=0mu 256\times256$ with all post-processing techniques.
	Accuracy for some classes is significantly lower than for others due to the fact that
	these classes contain smaller number of samples in train and validation datasets.
	For short sleeve outwear class it's 543 and 142 samples in train and validation subsets respectively,
	for sling class -- 1985 and 322. For comparison, vest dress contains 17949 and 3352 samples.}
	\label{tab:mdla-by-class-accuracy}
\end{table}

\section{Conclusion}\label{sec:conclusion}

This new approach is proposed as an adaptation of CenterNet~\cite{CenterNet} for clothing landmark estimation tasks.
The accuracy is comparable to state-of-the-art was achieved on the DeepFashion2 dataset by applying several
post-processing techniques as well as the semantic keypoint grouping approach: clothing
detection hit 0.737~$\mathit{mAP}$ and clothing landmark estimation reached 0.591~$\mathit{mAP}$.
Performance runs at 56 ms per image (more than 17 FPS) on a Huawei P40 Pro for mDLA-34 $\medmuskip=0mu 256\times256$,
and yields considerably high accuracy (0.445~$\mathit{mAP}_{pt}$ and 0.652~$\mathit{mAP}_{box}$) for clothing
detection tasks.

{\small
\bibliographystyle{ieee_fullname}
\bibliography{egpaper}

\begin{thebibliography}{10}\itemsep=-1pt

\bibitem{chen2019hybrid}
Kai Chen, Jiangmiao Pang, Jiaqi Wang, Yu Xiong, Xiaoxiao Li, Shuyang Sun,
  Wansen Feng, Ziwei Liu, Jianping Shi, Wanli Ouyang, et~al.
\newblock Hybrid task cascade for instance segmentation.
\newblock In {\em Proceedings of the IEEE conference on computer vision and
  pattern recognition}, pages 4974--4983, 2019.

\bibitem{Chen_2019_ICCV}
Ming Chen, Yingjie Qin, Lizhe Qi, and Yunquan Sun.
\newblock Improving fashion landmark detection by dual attention feature
  enhancement.
\newblock In {\em The IEEE International Conference on Computer Vision (ICCV)
  Workshops}, Oct 2019.

\bibitem{DCN}
Jifeng Dai, Haozhi Qi, Yuwen Xiong, Yi Li, Guodong Zhang, Han Hu, and Yichen
  Wei.
\newblock Deformable convolutional networks.
\newblock {\em CoRR}, abs/1703.06211, 2017.

\bibitem{DeepFashion2}
Yuying Ge, Ruimao Zhang, Xiaogang Wang, Xiaoou Tang, and Ping Luo.
\newblock {DeepFashion2}: A versatile benchmark for detection, pose estimation,
  segmentation and re-identification of clothing images.
\newblock In {\em Proceedings of the IEEE Conference on Computer Vision and
  Pattern Recognition}, pages 5337--5345, 2019.

\bibitem{MaskRCNN}
Kaiming He, Georgia Gkioxari, Piotr Doll{\'{a}}r, and Ross~B. Girshick.
\newblock Mask {R-CNN}.
\newblock {\em CoRR}, abs/1703.06870, 2017.

\bibitem{He_2020_CVPR}
Yisheng He, Wei Sun, Haibin Huang, Jianran Liu, Haoqiang Fan, and Jian Sun.
\newblock Pvn3d: A deep point-wise 3d keypoints voting network for 6dof pose
  estimation.
\newblock In {\em Proceedings of the IEEE/CVF Conference on Computer Vision and
  Pattern Recognition (CVPR)}, June 2020.

\bibitem{alibaba2020mnn}
Xiaotang Jiang, Huan Wang, Yiliu Chen, Ziqi Wu, Lichuan Wang, Bin Zou, Yafeng
  Yang, Zongyang Cui, Yu Cai, Tianhang Yu, Chengfei Lv, and Zhihua Wu.
\newblock Mnn: A universal and efficient inference engine.
\newblock In {\em MLSys}, 2020.

\bibitem{Kraus_2019}
Florian Kraus and Klaus Dietmayer.
\newblock Uncertainty estimation in one-stage object detection.
\newblock {\em 2019 IEEE Intelligent Transportation Systems Conference (ITSC)},
  Oct 2019.

\bibitem{law2018cornernet}
Hei Law and Jia Deng.
\newblock Cornernet: Detecting objects as paired keypoints.
\newblock In {\em Proceedings of the European Conference on Computer Vision
  (ECCV)}, pages 734--750, 2018.

\bibitem{Lin2020AggregationAF}
Tzu-Heng Lin.
\newblock Aggregation and finetuning for clothes landmark detection.
\newblock {\em ArXiv}, abs/2005.00419, 2020.

\bibitem{lin2017feature}
Tsung-Yi Lin, Piotr Doll{\'a}r, Ross Girshick, Kaiming He, Bharath Hariharan,
  and Serge Belongie.
\newblock Feature pyramid networks for object detection.
\newblock In {\em Proceedings of the IEEE conference on computer vision and
  pattern recognition}, pages 2117--2125, 2017.

\bibitem{liu2016ssd}
Wei Liu, Dragomir Anguelov, Dumitru Erhan, Christian Szegedy, Scott Reed,
  Cheng-Yang Fu, and Alexander~C Berg.
\newblock Ssd: Single shot multibox detector.
\newblock In {\em European conference on computer vision}, pages 21--37.
  Springer, 2016.

\bibitem{liu2016deepfashion}
Ziwei Liu, Ping Luo, Shi Qiu, Xiaogang Wang, and Xiaoou Tang.
\newblock Deepfashion: Powering robust clothes recognition and retrieval with
  rich annotations.
\newblock In {\em Proceedings of the IEEE conference on computer vision and
  pattern recognition}, pages 1096--1104, 2016.

\bibitem{miller2017dropout}
Dimity Miller, Lachlan Nicholson, Feras Dayoub, and Niko Sünderhauf.
\newblock Dropout sampling for robust object detection in open-set conditions,
  2017.

\bibitem{moon2018posefix}
Gyeongsik Moon, Ju~Yong Chang, and Kyoung~Mu Lee.
\newblock Posefix: Model-agnostic general human pose refinement network, 2018.

\bibitem{alej2016stacked}
Alejandro Newell, Kaiyu Yang, and Jia Deng.
\newblock Stacked hourglass networks for human pose estimation, 2016.

\bibitem{redmon2016you}
Joseph Redmon, Santosh Divvala, Ross Girshick, and Ali Farhadi.
\newblock You only look once: Unified, real-time object detection.
\newblock In {\em Proceedings of the IEEE conference on computer vision and
  pattern recognition}, pages 779--788, 2016.

\bibitem{EfficientGrouping}
Alexey Sidnev, Ekaterina Krasikova, and Maxim Kazakov.
\newblock Efficient grouping for keypoint detection.
\newblock {\em CoRR}, abs/2010.12390, 2020.

\bibitem{DeepMark}
Alexey Sidnev, Alexey Trushkov, Maxim Kazakov, Ivan Korolev, and Vladislav
  Sorokin.
\newblock Deepmark: One-shot clothing detection.
\newblock In {\em The IEEE International Conference on Computer Vision (ICCV)
  Workshops}, Oct 2019.

\bibitem{sun2019deep}
Ke Sun, Bin Xiao, Dong Liu, and Jingdong Wang.
\newblock Deep high-resolution representation learning for human pose
  estimation.
\newblock In {\em Proceedings of the IEEE conference on computer vision and
  pattern recognition}, pages 5693--5703, 2019.

\bibitem{Tian_2019_ICCV}
Zhi Tian, Chunhua Shen, Hao Chen, and Tong He.
\newblock Fcos: Fully convolutional one-stage object detection.
\newblock In {\em Proceedings of the IEEE/CVF International Conference on
  Computer Vision (ICCV)}, October 2019.

\bibitem{TruongLe2018UncertaintyEF}
Michael Truong-Le, Frederik Diehl, Thomas Brunner, and Alois Knoll.
\newblock Uncertainty estimation for deep neural object detectors in
  safety-critical applications.
\newblock {\em 2018 21st International Conference on Intelligent Transportation
  Systems (ITSC)}, pages 3873--3878, 2018.

\bibitem{DARK}
Feng Zhang, Xiatian Zhu, Hanbin Dai, Mao Ye, and Ce Zhu.
\newblock Distribution-aware coordinate representation for human pose
  estimation.
\newblock {\em arXiv preprint arXiv:1910.06278}, 2019.

\bibitem{Zhang_2018_ECCV}
Shifeng Zhang, Longyin Wen, Xiao Bian, Zhen Lei, and Stan~Z. Li.
\newblock Occlusion-aware r-cnn: Detecting pedestrians in a crowd.
\newblock In {\em Proceedings of the European Conference on Computer Vision
  (ECCV)}, September 2018.

\bibitem{zhang2014facial}
Zhanpeng Zhang, Ping Luo, Chen~Change Loy, and Xiaoou Tang.
\newblock Facial landmark detection by deep multi-task learning.
\newblock In {\em European conference on computer vision}, pages 94--108.
  Springer, 2014.

\bibitem{CenterNet}
Xingyi Zhou, Dequan Wang, and Philipp Kr{\"a}henb{\"u}hl.
\newblock Objects as points.
\newblock {\em arXiv preprint arXiv:1904.07850}, 2019.

\bibitem{Zhou_2019_CVPR}
Xingyi Zhou, Jiacheng Zhuo, and Philipp Krahenbuhl.
\newblock Bottom-up object detection by grouping extreme and center points.
\newblock In {\em Proceedings of the IEEE/CVF Conference on Computer Vision and
  Pattern Recognition (CVPR)}, June 2019.

\end{thebibliography}
}

\end{document}